# Video Detector: A Dual-Phase Vision-Based System for Real-Time Traffic Intersection Control and Intelligent Transportation Analysis

**Mustafa Fatih Şen [1], Halûk Gümüşkaya [2,\*], and Şenol Pazar [1,3]**

[1] Ankageo Co. Ltd., Yıldız Technical University, İkitelli Technopark, 34220 Istanbul, Türkiye
[2] Department of Computer Engineering, Istanbul Arel University, 34537 Istanbul, Türkiye
[3] Department of Electrical and Electronics Engineering, Biruni University, 34015 Istanbul, Türkiye
\* Correspondence: halukgumuskaya@arel.edu.tr

**Abstract:** Urban traffic management increasingly requires intelligent sensing systems capable of adapting to dynamic traffic conditions without costly infrastructure modifications. Vision-based vehicle detection has therefore become a key technology for modern intelligent transportation systems. This study presents Video Detector (VD), a dual-phase vision-based traffic intersection management system designed as a flexible and cost-effective alternative to traditional inductive loop detectors. The framework integrates a real-time module (VD-RT) for intersection control with an offline analytical module (VD-Offline) for detailed traffic behavior analysis. Three system configurations were implemented using SSD Inception v2, Faster R-CNN Inception v2, and CenterNet ResNet-50 V1 FPN, trained on datasets totaling 108,000 annotated images across 6–10 vehicle classes. Experimental results show detection performance of up to 90% test accuracy and 29.5 mAP@0.5, while maintaining real-time throughput of 37 FPS on HD video streams. Field deployments conducted in collaboration with Istanbul IT and Smart City Technologies Inc. (ISBAK) demonstrate stable operation under diverse environmental conditions. The system supports virtual loop detection, vehicle counting, multi-object tracking, queue estimation, speed analysis, and multiclass vehicle classification, enabling comprehensive intersection monitoring without the need for embedded road sensors. The annotated dataset and training pipeline are publicly released to support reproducibility. These results indicate that the proposed framework provides a scalable and deployable vision-based solution for intelligent transportation systems and smart-city traffic management.

**Keywords:** vehicle detection; intelligent transportation systems; traffic monitoring & management; computer vision

## 1. Introduction

Urban traffic management increasingly requires intelligent sensing systems capable of adapting to dynamic traffic conditions without costly infrastructure modifications. Vision-based vehicle detection has therefore emerged as a key technology for intelligent transportation systems. Traditional traffic control systems operating on fixed-timing schedules fail to adapt to real-time traffic variations, causing increased congestion and inefficient infrastructure usage [1–3]. These systems cannot account for critical variables such as varying traffic densities, peak flow periods, or situations where vehicles wait unnecessarily at signals [1].

Deep learning and computer vision technologies offer promising solutions for intelligent traffic management. Recent studies using Faster R-CNN have demonstrated strong detection performance in traffic management applications. Chaudhuri [4] reported high accuracy through adaptive background modeling with shadow compensation, while Abbas et al. [5] achieved 95.7% detection accuracy on CCTV streams for intelligent traffic light management. These capabilities enable sophisticated traffic analytics essential for optimizing signal timing and reducing congestion [6, 7, 8].

The integration of IoT technologies with machine learning enhances intelligent transportation systems through cloud computing, edge processing, and wireless networks, creating scalable solutions adaptable to varying traffic conditions [9, 10, 11]. Vision-based systems offer significant advantages over traditional sensors, including lower deployment costs and analytics from single camera installations [3, 12].

This paper presents *Video Detector*, a comprehensive deep learning-based traffic intersection management system, addressing limitations of conventional approaches. Our solution leverages advanced computer vision for virtual loop counting, vehicle tracking, queue analysis, and multi-class classification, enabling cost-effective deployment of intelligent traffic intersection management for smart cities.

*1.2. Related Work*

Traditional traffic monitoring systems have relied on inductive loop detectors, radar, and infrared sensors. While loop detectors provide reliable measurements of traffic volume and occupancy, they suffer from high installation and maintenance costs and are limited in their ability to classify vehicle types or adapt to changing traffic patterns [12,13]. Early computer vision–based approaches attempted to address these limitations but were often sensitive to shadows, weather conditions, and illumination changes, constraining their practical deployment [12].

Recent work has increasingly shifted toward deep learning–based object detection models, which offer improved robustness and accuracy under diverse traffic and environmental conditions. Deep learning approaches to vehicle detection can be broadly categorized into two-stage and single-stage architectures, which differ primarily in their trade-offs between detection accuracy and inference speed. Two-stage detectors, such as Faster R-CNN, first generate region proposals and then refine classifications, typically achieving higher accuracy at the expense of computational efficiency. Chaudhuri [4] demonstrated the effectiveness of Faster R-CNN for adaptive traffic management using background modeling and shadow compensation, while Abbas et al. [5] reported high detection accuracy for intelligent traffic light control using CCTV streams. Wang et al. [14] further improved vehicle classification performance by enhancing region proposal mechanisms.

Single-stage detectors perform object detection in a single forward pass, directly predicting bounding boxes and class probabilities. These models are generally faster and therefore well suited for real-time traffic monitoring applications. YOLO-based architectures have been widely adopted in this context due to their high frame rates. Song et al. [7] applied YOLOv3 for highway vehicle counting, while more recent studies have explored advanced variants such as YOLOv8 for smart city applications [15,16] and YOLOv9 for dense urban traffic scenarios [17]. Additional extensions, including attention-based designs such as TSD-YOLO, have been proposed to improve detection of small or challenging traffic objects [18].

Beyond single-model detectors, several studies have investigated hybrid and multi-stage pipelines that combine complementary components to balance accuracy and efficiency. Mittal et al. [19] introduced EnsembleNet, which integrates Faster R-CNN and YOLO to leverage the strengths of both architectures. Other approaches have incorporated segmentation, tracking, and optical character recognition modules to enable richer traffic analysis. For example, Li et al. [6] proposed a multi-stage framework combining segmentation, tracking, and license plate recognition to achieve robust performance in high-density traffic environments.

In addition to ground-based camera systems, aerial and UAV-based approaches have been explored for large-scale and mobile traffic monitoring. Singh et al. [11] integrated YOLOv8 with Transformer-based convolutional networks to analyze aerial imagery, while Zhu et al. [20] demonstrated real-time urban traffic analysis on resource-constrained devices using UAV video streams. For real-time processing, several studies have focused on optimizing inference efficiency through model design and hardware acceleration. TensorRT-based optimizations and related techniques have been shown to significantly improve inference speed while maintaining acceptable accuracy [21–24].

While accurate vehicle detection is a fundamental requirement, effective traffic management systems extend beyond detection to include traffic control, congestion analysis, and decision-making mechanisms. IoT-enabled adaptive traffic management systems have been proposed to integrate detection outputs with signal control and congestion prediction [8–10]. Gandhi et al. [2] demonstrated AI-driven traffic signal optimization, while Kanungo et al. [3] proposed video-based traffic density estimation methods that outperform traditional hardware sensors. Hybrid traffic forecasting and congestion analysis models have also been explored to support long-term planning and operational decision-making [25, 26].

Collectively, existing studies highlight important trade-offs among detection accuracy, processing speed, and deployment constraints. However, many prior works focus on isolated components—such as detection, tracking, or control—without providing a unified, deployable framework that integrates real-time signal control with comprehensive offline analytics. In addition, practical considerations such as cross-junction generalization, cost-effectiveness, and large-scale municipal deployment are often underexplored. These gaps motivate the development of integrated, scalable traffic management systems that bridge the divide between algorithmic performance and real-world operational requirements.

Despite the rich body of research on intelligent traffic management, several critical gaps remain unaddressed in existing studies. First, most deep-learning–based traffic detection systems focus solely on object detection or tracking and do not integrate real-time traffic signal control with comprehensive offline analytical capabilities within a unified framework. Second, while many works explore individual detection models—such as Faster R-CNN, YOLO variants, or hybrid approaches—few studies provide a scalable architecture capable of continuous learning across multiple intersections, addressing model generalization challenges encountered in heterogeneous urban environments. Third, prior research rarely evaluates cost-effectiveness in practical deployments, frequently relying on high-end hardware, subscription-based commercial systems, or laboratory-only setups, limiting municipal scalability. Finally, although several studies incorporate advanced analytics such as OD matrices or trajectory extraction, these often require multi-camera installations or assume ideal field-of-view conditions without providing deployable solutions for real-world constraints.

The proposed Video Detector system directly addresses these gaps by introducing a dual-phase architecture that unifies real-time intersection control (VD-RT) with advanced post-event analysis (VD-OffLine) in a single, flexible platform. The system leverages a continuously expanding training dataset sourced from diverse intersections to enhance model generalization, thereby overcoming the limitations of junction-specific models common in prior work. Furthermore, by prioritizing deployment feasibility—through cost-effective hardware options, simplified installation, and collaboration with a major municipal traffic authority (ISBAK)—the study demonstrates practical scalability beyond controlled research environments. While many works address parts of this pipeline, few provide a unified, deployable system that combines real-time signal control, offline analytics, and explicit cost evaluation as in the VD framework. This work therefore aims to bridge the gap between high-accuracy deep-learning–based vehicle detection and practical, scalable traffic management deployment by introducing a unified framework that combines real-time intersection control with comprehensive offline traffic analytics.

*1.3. Contributions*

This study presents a deployable, vision-based traffic intersection management system with explicit technical design choices targeting scalability, robustness, and operational feasibility. The main contributions are summarized as follows:

1. **Dual-phase system architecture:** A unified framework is introduced that combines real-time intersection control (VD-RT) with offline, high-precision traffic analytics (VD-OffLine). This separation enables stable real-time operation while supporting detailed post-event analysis within a single end-to-end system.
2. **Scalable unified training strategy for cross-junction generalization:** A single-model training approach is adopted in which the dataset continuously expands as new intersections are added, avoiding junction-specific models. This strategy improves cross-junction generalization, reduces model maintenance overhead, and supports scalable municipal deployment.
3. **Explicit class-imbalance mitigation via controlled data augmentation:** A targeted augmentation pipeline is developed to address severe class imbalance by selectively augmenting minority classes while excluding the majority class. This approach reduces the dominant class ratio from approximately 67% to 40% without full class equalization, improving robustness while limiting overfitting.
4. **Performance-driven model progression across system versions:** Multiple object detection backbones (SSD Inception v2, Faster R-CNN Inception v2, and CenterNet ResNet-50) are implemented and benchmarked. The system evolution reflects deployment constraints, transitioning from early real-time stability to a TensorFlow-2–based CenterNet configuration that provides the best speed–accuracy trade-off (90% test accuracy, 37 FPS) for high-precision analysis.
5. **Reproducible quantitative evaluation with public data release:** System performance is evaluated using standard detection metrics, including accuracy (API-consistent definition), mAP@0.5, inference speed, and precision–recall curves. The annotated junction dataset and training pipeline are publicly released to support reproducibility and comparative evaluation.
6. **Deployment and economic validation through municipal collaboration:** The system is validated through real-world deployments conducted in collaboration with Istanbul IT and Smart City Technologies Inc. (ISBAK). A comparative 5-year total cost of ownership (TCO) analysis demonstrates the economic advantages of the proposed approach over inductive loop detectors and commercial vision-based platforms. The practical viability of transitioning research concepts to deployed solutions in Istanbul provides valuable technical and bureaucratic implementation insights for future intelligent transportation system deployments.

## 2. Dual-Mode Architecture of the Video Detector System

The Video Detector (VD) traffic management system has three versions: VD-Real Time (VD-RT), VD-OffLine-v1 and VD-OffLine-v2. The specifications of the basic hardware components, operating systems and deep learning platform TensorFlow used in all VD versions are given in Table 1.

**Table 1.** The VD hardware and software specifications.

| | |
|---|---|
| **CPU** | Intel Core i7-7500U @3.5 GHz (VD-RT) <br> Intel Core i7-9700K @3.6 GHz (VD-OffLine-v1) and Intel Core i7-12700K @3.6 GHz (VD-OffLine-v2) |
| **GPU** | NVIDIA GeForce 1050ti (VD-RT) <br> NVIDIA GeForce GTX 1660 Ti (VD-OffLine-v1) and NVIDIA GeForce RTX 3090 (VD-OffLine-v2) |
| **Memory** | 16 GB (VD-RT / VD-OffLine-v1) and 64 GB (VD-OffLine-v2) |
| **Operating System** | Windows 10 (64 bit) (VD-RT) and Linux Ubuntu (VD-OffLine-v1 / VD-OffLine-v2) |
| **TensorFlow** | TensorFlow 1.13 (VD-RT)  /  TensorFlow 2.14 (VD-OffLine-v1 / VD-OffLine-v2) |

*2.1. Real-Time Subsystem (VD-RT): Hardware–Software Architecture*

The VD-RT system is a real-time traffic solution designed to improve intersection control through intelligent video analysis. The simplified hardware and software architectures of VD-RT are shown in Figure 1 and Figure 2. The VD-RT system can be decomposed into 4 main layers: *data acquisition*, *processing*, *communication* and *control*.

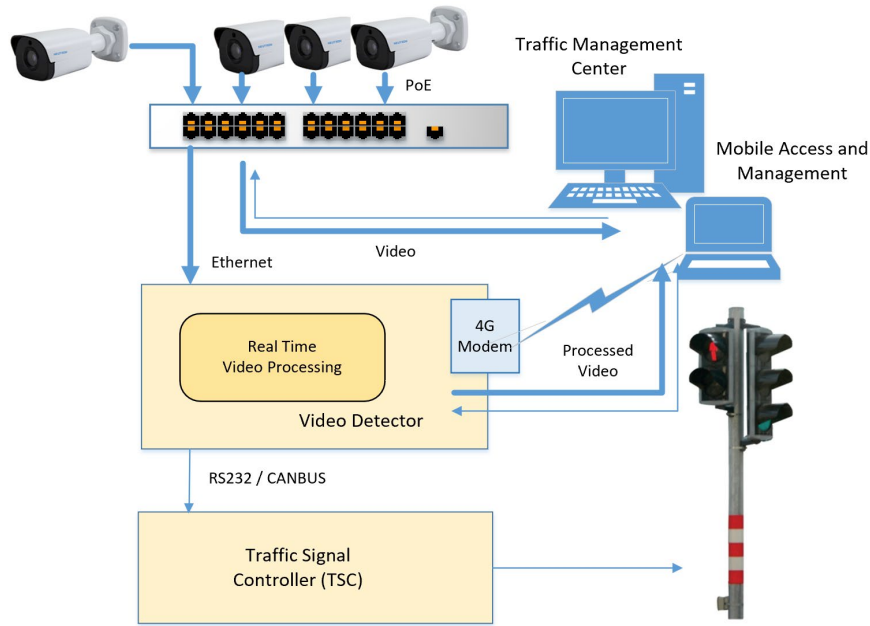

**Figure 1.** The hardware architecture of VD-RT: Traffic monitoring and traffic lights control.

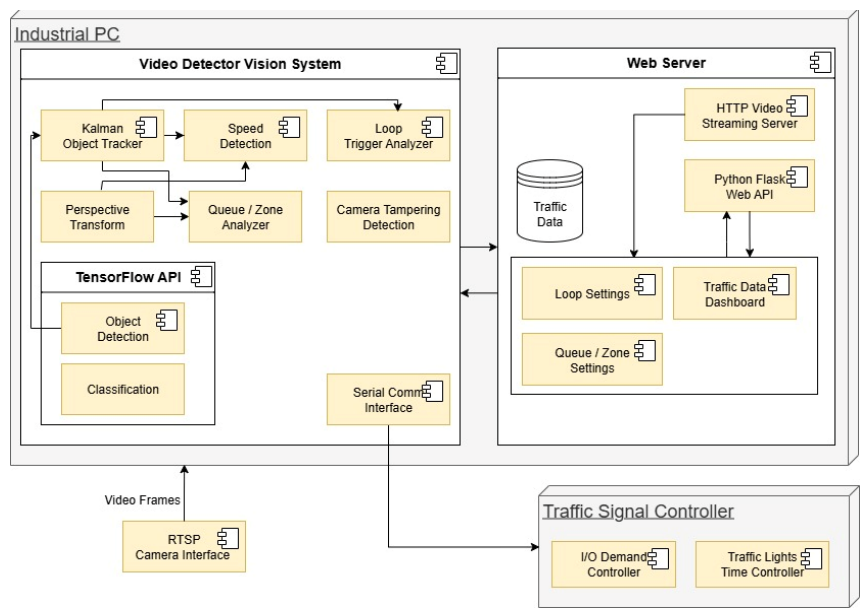

**Figure 2.** The software architecture of VD-RT.

The *data acquisition* layer employs Power over Ethernet (PoE) IP cameras strategically positioned at intersections to monitor one-way traffic flow, connected through a robust network switch that provides both power and data transmission via secure underground cables. At a traffic junction, a VD-RT unit is connected to cameras via Ethernet, continuously receiving and processing video streams. Within the application interface, operators can define virtual loops that function as digital equivalents of traditional metal detectors: whenever a vehicle passes over a loop, the system generates a presence signal. These signals are then transmitted to the Traffic Signal Controller (TSC), which adjusts the signal phases accordingly.

The *processing layer* centers around an industrial PC equipped with a specialized NVIDIA GeForce graphics card capable of analyzing video streams from up to 4 cameras simultaneously. A web server runs on the PC and is connected to the Internet. The core intelligence is powered by the SSD Inception v2 deep learning model, trained on a dataset of 62,000 samples to recognize and classify 6 distinct vehicle types.

Leveraging recent advancements in NVIDIA graphics technology, we migrated to compact, cost-efficient Jetson-based boards to substantially lower hardware expenses. Earlier prototypes relied on industrial PCs because Jetson-class devices were not yet mature enough to support stable real-time inference at the time of initial development. With the increasing performance and reliability of modern embedded GPUs, the VD-RT system can now operate on significantly more affordable edge hardware, further improving its overall cost-effectiveness.

The software architecture as shown in Figure 2 integrates multiple vision components such as vehicle detection, tracking, virtual loop activation, queue-occupancy estimation, OD-matrix generation, speed calculation, and long-term traffic behavior analysis. In the following section, these components are presented.

*2.2. Tracking Module Integration with Virtual Loops and Signal Control*

The VD-RT system is designed to perform real-time vehicle detection, tracking, and virtual loop activation for adaptive intersection control. Video streams captured by PoE IP cameras are processed by different software components of the Video Detector Vision System, where the SSD Inception v2 model performs object detection. The model outputs bounding box coordinates and class labels for each frame, forming the primary input for the tracking module.

The VD-RT system was developed and deployed during an early phase of the project when SSD Inception v2 was the most stable real-time model supported in the TensorFlow 1.x framework. Due to regulatory and operational constraints associated with updating traffic controller systems, the deployed model could not be replaced despite later advancements. Although SSD Inception v2 currently satisfies the functional requirements of the VD-RT system, recent lightweight detectors such as YOLOv8-n demonstrate significantly improved computational efficiency, enabling comparable or higher detection accuracy with lower latency. Consequently, migrating to a modern architecture like YOLOv8-n would potentially allow the system to operate on lower-end hardware, leading to reduced deployment costs and improved scalability without sacrificing real-time performance.

To maintain consistent vehicle trajectories across frames, the VD-RT system integrates a multi-object tracking module that operates on the detection results. This tracking component is essential for several system functions, including queue-length estimation, speed calculation, and reliable activation of virtual loops.

To ensure temporal continuity in vehicle trajectories within the VD-RT system, we employ a Kalman filter–based multi-object tracking approach, a widely adopted method in vision-based traffic monitoring due to its computational efficiency and robustness under partial occlusion [28, 29]. Detection outputs from the SSD Inception v2 model provide bounding box coordinates for each frame, which are passed to the tracker to maintain consistent object identities over time. Our implementation uses a six-dimensional state vector modeling the object center position, velocity, and bounding box size, while new detections are associated with predicted states through an IoU-based matching strategy. This pipeline produces stable tracking IDs even during temporary detection gaps and yields continuous trajectories that are stored in a PostgreSQL database. These trajectories form the foundation for system-specific analytics—including virtual loop activation, queue-occupancy estimation, OD-matrix generation, speed calculation, and long-term traffic behavior analysis— moving beyond frame-by-frame detections toward a comprehensive understanding of intersection dynamics.

The stable tracking identifiers generated by the Kalman filter–based module enable reliable activation of user-defined virtual loops, which function as software equivalents of inductive loop detectors. Loop activation events are transmitted to the Traffic Signal Controller (TSC), allowing adaptive signal phase control based on real-time vehicle presence. In parallel, queue occupancy is estimated by monitoring tracked objects within predefined regions, enabling congestion-aware intersection management. These outputs are continuously logged and made available to the central management system for monitoring and retrospective analysis.

*2.3. Real-Time Vehicle Detection, Tracking, and Virtual Loop Activation for Intersection Control*

The VD-RT's defining feature is its real-time operation, enabling instantaneous video stream processing for vehicle detection, tracking, and virtual loop activation. Users can define virtual loops at strategic locations within the camera's field of view, creating customizable monitoring zones that replace traditional inductive loop detectors.

The VD-RT's multi-stage process enables continuous and robust tracking of vehicles from entrance to exit, maintaining consistent IDs even under brief occlusions or detection interruptions. The resulting tracking numbers support higher-level operations such as virtual loop creation, queue-length monitoring, and speed estimation, all configurable remotely through the web server. Users can monitor processed video streams in real-time at the traffic management center and dynamically adjust detection parameters, including virtual loop placement and sensitivity thresholds, as illustrated in Figure 3.

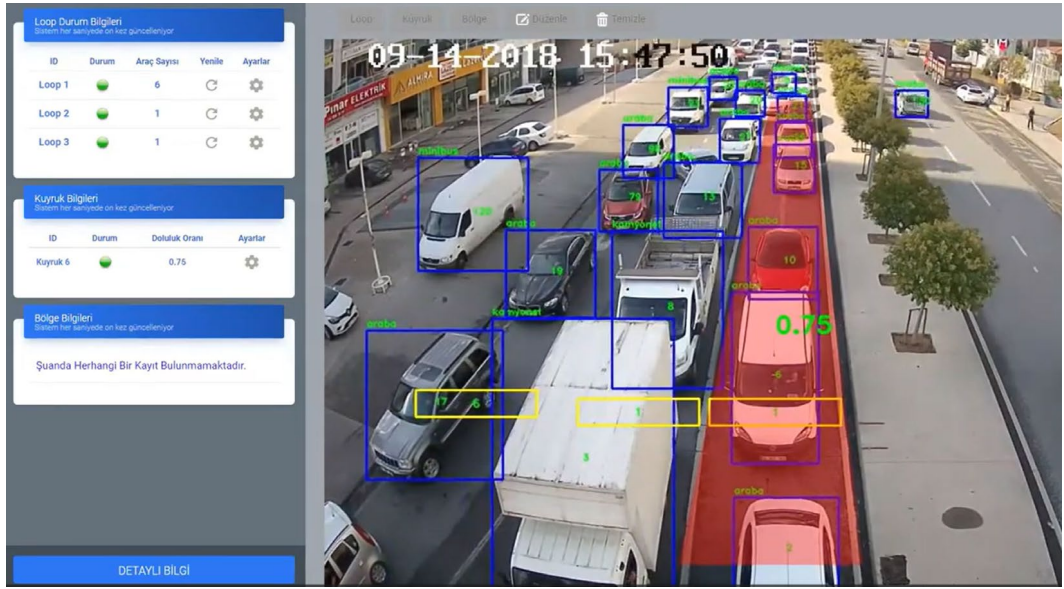

**Figure 3.** Example of VD-RT real-time operation under diverse environmental conditions, demonstrating vehicle detection, tracking, virtual loop activation, and queue density estimation.

The figure presents the VD-RT system operating under diverse environmental conditions such as nighttime, rain, and snow, demonstrating robust performance. It showcases simultaneous multi-vehicle detection and tracking, virtual loop triggering, and queue density estimation within a single camera view. These integrated capabilities enable adaptive intersection control without the need for physical, road-embedded sensing infrastructure.

The *communication layer* utilizes a 4G router USB modem to enable remote access, monitoring and data transmission to a central traffic management center. Remote access capabilities allow traffic engineers to monitor multiple intersections from a central location, receiving real-time data about traffic conditions and system performance. It is also possible to access the Video Detector system remotely, and users can define virtual loops anywhere in the video. The user can control the location and sensitivity of virtual loops. It is also possible to create queue calculation areas as shown in Figure 3.

The *control layer* generates signals that integrate directly with existing TSCs. The integration with existing infrastructure ensures seamless deployment without requiring extensive modifications to current traffic control systems. The system continuously monitors real-time traffic flow and sends the vehicle data to TSC units.

*2.4. Real-Time Traffic Statistics*

The vehicle detection, video processing, and traffic light control capabilities of the VD-RT system were successfully tested at different locations of Istanbul traffic and weather conditions. Real-time traffic statistics can be generated as different reports at the traffic management center or on a mobile management computer. A sample window of traffic statistics is given in Figure 4. This dashboard has graphs that show the vehicle density over hours and days. The user can select a camera corresponding to a specific intersection lane and filter results by time. The available management functions of all Video Detector versions are given and explained in Section 6 and Table 3.

The VD-RT system generates real-time and historical traffic statistics, including vehicle counts by class, queue occupancy, and temporal traffic density trends. These statistics are aggregated per detector and per time interval, allowing traffic engineers to assess intersection performance, identify peak congestion periods, and evaluate signal timing effectiveness. The reporting interface supports filtering by camera, detector, and time range, enabling flexible exploration of traffic patterns without manual data processing.

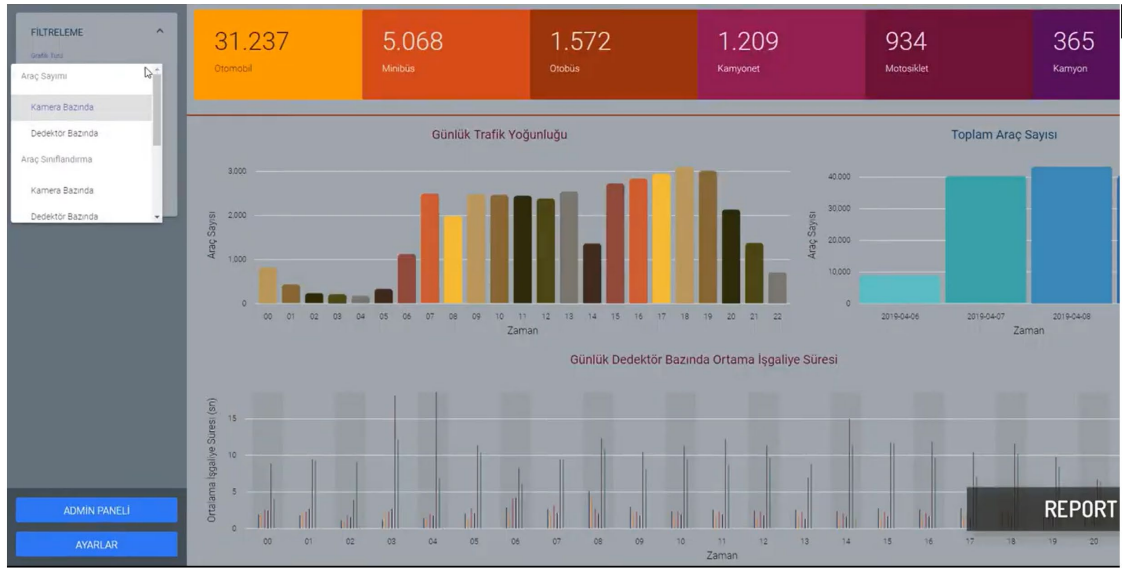

**Figure 4.** Real-time traffic statistics: Vehicle classes/numbers, occupancy and other information.

## 3. Offline Subsystem (VD-OffLine-v1): Operator-Assisted Analysis

The VD-OffLine-v1 addresses the growing need for comprehensive traffic analysis in scenarios where real-time control is not the primary objective, instead focusing on detailed post-processing analysis of traffic patterns. With this system, users can upload videos and obtain insights such as average speed, OD matrix, average occupancy time, peak times, and so on.

The VD-OffLine-v1 workflow was built to support user accessibility while allowing operator-assisted calibration. Users create project-specific workspaces and upload their recorded videos with basic timing information. The system then forwards each job to an operator, who reviews the footage and manually positioned virtual detection loops to ensure accurate lane alignment and minimize counting errors. An example of this operator calibration interface is shown in Figure 5, where virtual loops are adjusted over the video to prepare the data for automated analysis.

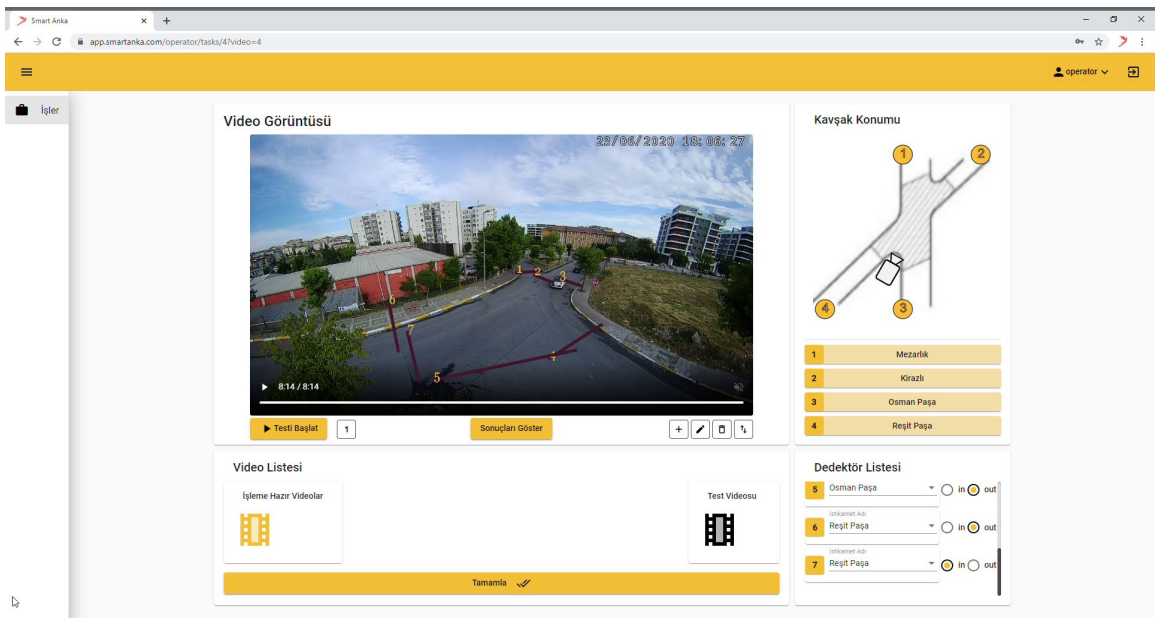

**Figure 5.** The operator interface for defining and optimizing virtual loop placement.

After the operator completes the loop calibration, the user is notified and can access the generated reports. A sample reporting interface is shown in Figure 6. All extracted trajectories and counts are stored in a PostgreSQL database, from which the system produces analytical outputs such as vehicle counts, turning movements, and temporal traffic patterns. These reports support traffic engineers in evaluating intersection performance and identifying congestion trends. VD-OffLine-v1 thus offers a more accurate and consistent alternative to manual traffic counting, although its operator-dependent workflow motivated the development of the fully automated VD-OffLine-v2.

VD-OffLine-v1 employs an operator-assisted calibration workflow in which virtual detection loops are manually aligned with traffic lanes prior to automated analysis. This calibration step improves counting accuracy and trajectory extraction, particularly in complex junction geometries. Once calibrated, the system automatically extracts vehicle trajectories and generates analytical outputs such as turning movements, temporal traffic distributions, and aggregated vehicle counts.

As shown in Figure 6, the system provides access to various reporting capabilities including time-based vehicle counting, visualization of the Origin-Destination (OD) matrix which displays the number of vehicles for each movement from one direction to another, and other types of reports designed to satisfy the specific needs of traffic engineers.

Figure 6 presents a comprehensive traffic count summary generated by the VD-OffLine-v1 system. The reporting output combines time-series vehicle counts for entering and exiting flows with an Origin–Destination (OD) matrix that visualizes directional traffic movements between intersection approaches. Flow magnitudes are encoded by connection thickness, enabling rapid identification of dominant turning movements and congestion patterns. Aggregated vehicle counts by class are also provided, and the results can be exported for further analysis and archival purposes.

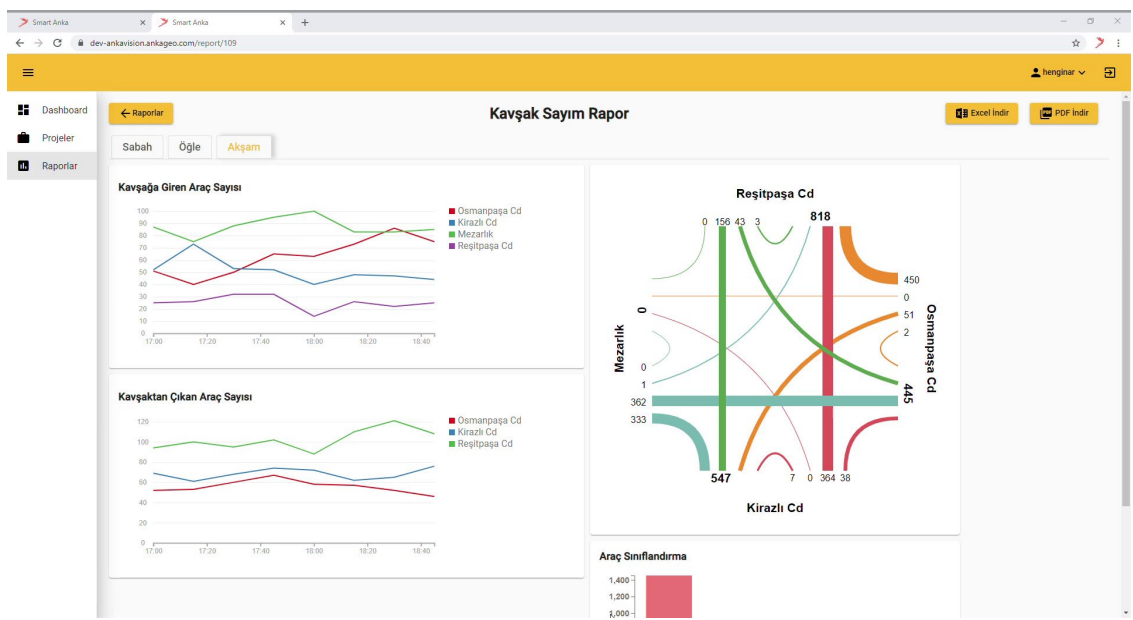

**Figure 6.** Example reporting output of the VD-OffLine-v1 system, combining time-series vehicle counts with Origin–Destination (OD) matrix visualization for directional traffic flow analysis.

## 4. Offline Subsystem (VD-OffLine-v2): Fully Automated Cloud-Based Analysis

VD-OffLine-v2 represents the most advanced stage of the system's evolution, incorporating state-of-the-art deep learning architectures and high-performance computing resources to deliver advanced analysis capabilities. This version implements dual-model architectures—Faster R-CNN Inception v2 and CenterNet ResNet-50 V1 FPN (512x512)—trained on the same 108,000-sample dataset to classify 10 vehicle categories.

The dual-model approach provides comprehensive analysis capabilities by leveraging the strengths of both architectures. By combining Faster R-CNN Inception v2 and CenterNet ResNet-50, the system benefits from complementary strengths: the former ensures precise detection and classification, while the latter provides faster inference suitable for real-time deployment. This design enables flexible adaptation depending on whether detection accuracy or processing speed is prioritized. This multi-model framework enables the system to adapt its analysis approach based on specific requirements, whether prioritizing detection speed or classification precision.

VD-OffLine-v2 operates within a cloud-based architecture, where video processing tasks such as detection, tracking, and trajectory extraction are executed on centralized servers rather than local machines. This architecture allows videos from multiple intersections to be uploaded and analyzed through a unified platform, reducing the need for users to maintain high-performance hardware on-site. The cloud environment also offers practical flexibility, as computational resources can adapt to varying workloads and larger datasets. Software updates and model improvements are deployed directly on the server, which helps minimize maintenance responsibilities for end users. Remote multi-user access further supports collaborative work among traffic engineers and planners who share project data and analysis outputs. As a result, the cloud-based design improves the system's accessibility and ease of use while supporting broader traffic analysis needs.

The system maintains the user-friendly workflow established in the previous version while adding enhanced analytical depth and accuracy. The same project-based approach allows users to upload traffic videos and define analysis parameters. After the video upload, the system autonomously analyzes the footage, extracts vehicle trajectories, and records them in a PostgreSQL database, removing the need for manual operator input.

The operator-based solution proved to be costly and time-consuming, so we developed VD-OffLine-v2 that provides the following steps for traffic data analysis:

**1.** User Interface and Video Upload: User uploads video files based on 3 categories:
- Project (corresponds to a junction) (Figure 9)
- Video group (videos captured sequentially)
- Video

**2.** Work creation: User uploads the videos and sets the timing information for them. Then, the user sends this job for processing. In this stage, the job is forwarded to our AI server for processing, where bounding boxes are created and trajectories are detected. Upon completion, a notification is sent to the user and the results are moved to the reporting section.

**3.** Report Generation: After video processing is completed, users create virtual loops and, based on these loops, generate OD matrices and other useful reports as shown in Figure 7. Then users go to reporting page and create their own reports based on the virtual loops they draw.

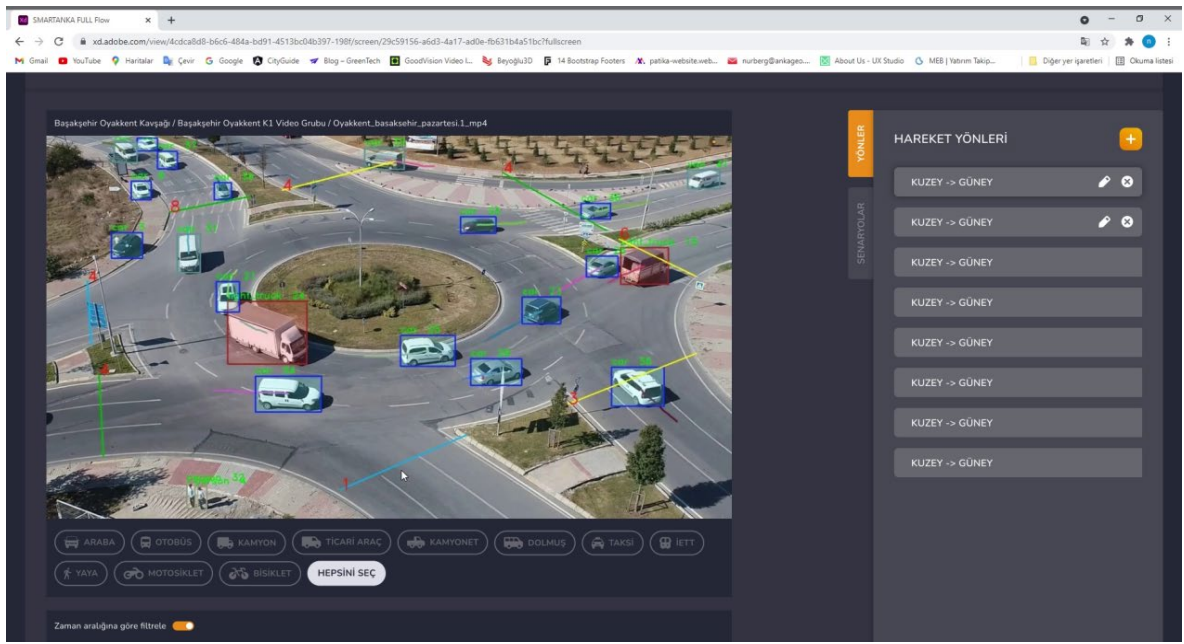

**Figure 7.** Automated traffic analysis workflow in the VD-OffLine-v2 system, illustrating trajectory extraction and user-defined movement analysis for OD matrix generation.

This enhanced video processing power enables more detailed analysis outputs, including advanced metrics such as vehicle speed estimation, acceleration patterns, and complex movement trajectory analysis. The system generates comprehensive reports that provide insights into traffic behavior patterns, peak-hour characteristics, and long-term trends that can inform infrastructure planning and traffic management strategies. It also creates an OD matrix for use in simulation programs. The improved accuracy and analytical depth make this version particularly suitable for research applications, detailed traffic studies, and high-stakes planning decisions where precision is paramount.

As shown in Figure 7, users can define the start and end directions for vehicle counting to create custom movement patterns. This feature allows traffic engineers to specify precise counting zones and directional flows according to their specific analysis requirements. Users have full control over determining the geographical bounds of counting areas, enabling them to focus on particular intersections, road segments, or traffic corridors of interest. Based on the created movements and defined counting parameters, users receive detailed counting reports that provide traffic flow data for their designated areas.

Figure 7 demonstrates the fully automated VD-OffLine-v2 workflow, in which vehicle trajectories are extracted without operator intervention and used to generate customized movement definitions and Origin–Destination (OD) matrices. This design enables scalable, cloud-based traffic analysis across multiple intersections while preserving analytical flexibility for research and planning applications. A modular reporting framework allows users to generate customized analytical dashboards by selecting relevant traffic metrics, supporting flexible post-processing without manual data manipulation.

## 5. Vision Models, Training Strategy, and Performance Evaluation

A summary of the Video Detector vision system is given in Table 2. It shows the models used by all three versions of Video Detector, training, validation and test accuracies, training times, dataset sizes, and vehicle classes.

**Table 2.** Video Detector versions, models, training, validation, and test accuracy values, dataset sizes, and vehicle classes.

| Video Detector | Model | Accuracy (%) | | | Training Time (h) | Dataset Size | Vehicle Classes |
|---|---|---|---|---|---|---|---|
| | | Training | Valid. | Test | | | |
| VD-RT | SSD Inception v2 | 89.6 | 87.4 | 86.80 | 48 | 62,000 | 6 [1] |
| VD-OffLine-v1 | SSD Inception v2 | 89.7 | 87.8 | 87.43 | 80 | 108,000 | 10 [2] |
| VD-OffLine-v2 | Faster R-CNN Incept. v2 | 91.2 | 90.04 | 89.20 | 80 | 108,000 | 10 [2] |
| | CenterNet ResNet-50 V1 FPN (512x512) | 93.24 | 91.21 | 90 | 36 | | |

[1] Car, Bus, Minibus, Motorcycle, Person, Truck,
[2] Car, Bus, Minibus, Motorcycle, Person, Taxi, Light Truck, Heavy Truck, Van, Bicycle

Accuracy in Table 2 denotes the proportion of correctly detected and correctly classified bounding boxes at the default IoU threshold of the TensorFlow Object Detection API and is reported for consistency with prior ITS studies. Standard object detection metrics (mAP@0.5) are reported separately for cross-model comparison.

VD-RT was the first version for real-time detection and video processing for 6 different vehicles using the SSD Inception v2 model. This first version has a simpler dataset setup, focusing on fewer samples and classes with no advanced preprocessing or augmentation techniques. The VD-OffLine-v1 system utilized the SSD Inception v2 model for offline processing, allowing for the classification of 10 vehicle classes. In the second offline version, VD-OffLine-v2, the Faster R-CNN Inception v2 and CenterNet ResNet-50 V1 FPN (512x512) were used for again 10 classes.

*5.1. Model Performance Comparison*

The VD-RT, using SSD Inception v2 on 62,000 samples and 6 classes, achieves 86.8% test accuracy with about 48 hours of training, making it suitable for real-time use. VD-OffLine-v1, trained on a larger dataset of 108,000 samples and 10 classes, slightly improves accuracy to 87.43% but requires about 80 hours. VD-OffLine-v2 applies more advanced models: Faster R-CNN Inception v2 reaches 89.2% test accuracy, while CenterNet ResNet-50 with FPN achieves the best result—90% test accuracy with the shortest training time of 36 hours. Overall, while SSD-based models are faster and lighter, the advanced architectures provide better accuracy and robustness. CenterNet ResNet-50 offers the best solution for accuracy and efficiency for offline processing.

The learning curves for the last model CenterNet ResNet-50 used in VD-OffLine-v2 are presented in Figure 8. The left part shows that both training and validation accuracy steadily increase with training steps, ultimately exceeding 0.90. The validation curve closely follows the training curve, indicating good generalization with only a small accuracy gap. In contrast, as shown in the right part of Figure 8, the training loss decreases rapidly to near zero, while the validation loss levels off at a higher value, revealing mild overfitting for this particular trace. Overall, the model achieves strong predictive performance, though techniques such as early stopping, dropout, or data augmentation could further improve robustness on unseen data.

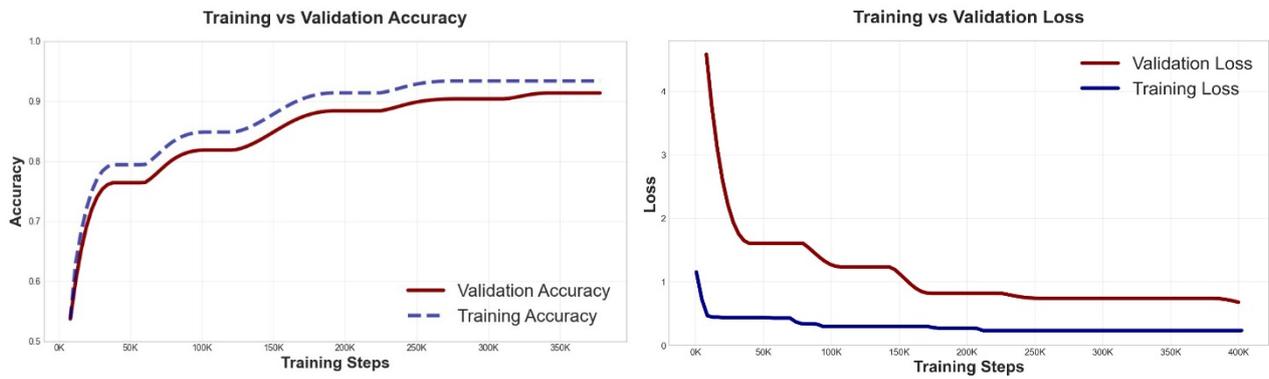

**Figure 8.** The learning curves of the CenterNet ResNet-50 model.

In this study, 'accuracy' refers to the proportion of correctly detected and correctly classified bounding boxes in the dataset, using the default IoU threshold defined by the TensorFlow Object Detection API. This metric is reported to ensure consistency with several referenced works that also present accuracy-based results. Precision–Recall analyses are additionally included in Figure 10 to provide a more detailed performance assessment.

In addition to the accuracy metric, we report mean Average Precision (mAP) to align with standard object detection evaluation practices and enable direct comparison with prior studies. In particular, mAP@0.5 is used as the primary benchmark, as it is the most commonly reported metric in traffic and intelligent transportation system (ITS) literature and reflects practical detection performance under operational IoU thresholds. While mAP@0.5:0.95 provides a stricter multi-threshold evaluation, mAP@0.5 remains more representative of real-time and deployment-oriented scenarios, where stable detection and classification are prioritized over marginal localization refinements. Accordingly, accuracy, mAP, and precision–recall curves are jointly reported to provide a comprehensive and transparent assessment of model performance.

The offline versions of Video Detector system leverage significantly larger datasets and incorporate advanced augmentation techniques and strategies to mitigate class imbalance. The introduction of random black masking further diversifies training data and reduces overfitting to dominant classes. These new improvements aim to create more robust models capable of handling diverse and noisy real-world scenarios.

Advanced data augmentation techniques form the backbone of this system's enhanced generalization capabilities. The training process incorporates random noise addition to improve robustness against various environmental conditions, color adjustments to handle different lighting scenarios throughout the day, and geometric transformations to account for varying camera angles and positions. Random black masking techniques are employed to reduce overfitting tendencies and ensure the model generalizes well across diverse traffic scenarios. Class imbalance mitigation strategies generate additional training data for underrepresented vehicle categories while carefully avoiding over-inflation of majority classes, resulting in a more balanced and versatile detection system.

By incrementally refining the models and adopting advanced architectures, the detection and tracking of vehicles at road junctions have been significantly improved. These enhancements contribute to a more accurate analysis of traffic patterns and more effective optimization of vehicle movements.

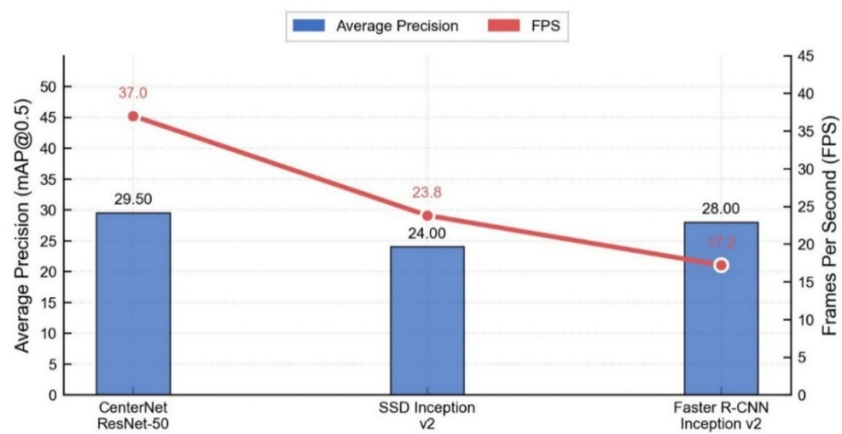

**Figure 9.** Comparison of models in terms of detection accuracy and processing speed on the COCO Dataset.

The performance comparison on the COCO dataset given in Figure 9 between three deep learning models demonstrates the trade-off between accuracy and processing speed in Video Detector systems. CenterNet ResNet-50 achieves the highest detection performance with 29.50 mAP@0.5 and the highest throughput of 37 FPS, followed by Faster R-CNN Inception v2 (28.00 mAP@0.5, 17.2 FPS) and SSD Inception v2 (24.00 mAP@0.5, 23.8 FPS). This analysis reveals that CenterNet ResNet-50 provides the optimal combination of detection accuracy and processing speed. Therefore, we switched to TensorFlow 2 models, which are more accurate and run faster.

*5.2. Training Framework and Datasets*

We used the TensorFlow Object Detection API [30] for all our models. It provides a range of pretrained models from the TensorFlow Model Garden, trained on COCO (Common Objects in Context), Open Images, Pascal VOC, KITTI, and other datasets, with support for fine-tuning on custom data. COCO offers 80 object categories with bounding boxes, segmentation masks, keypoints, and captions, while Open Images includes over 600 categories with broader and more fine-grained coverage. Pascal VOC provides high-quality annotations for 20 categories, and KITTI focuses on autonomous driving scenarios. These datasets collectively enable the API to support applications such as object detection, instance segmentation, human pose estimation, and image captioning, while maintaining flexibility for domain-specific customization.

*5.3. Data Preparation and Augmentation*

The dataset was constructed from multiple sources, with images captured by various cameras at different resolutions and sizes [31]. Annotations were conducted using the Computer Vision Annotation Tool (CVAT) platform under a multi-annotator work-flow with cross-review and controlled verification, based on predefined class definitions and annotation guidelines. This diversity in data sources enhances the model's ability to generalize across a wide range of scenarios and detect features accurately for all defined classes.

Deep learning models, especially those used for object detection, are highly sensitive to the quality, diversity, and balance of the training data. Achieving high detection accuracy requires a large and varied dataset. In this study, we utilized a dataset comprising 82,700 images, annotated with 275,530 labels across 10 distinct classes: car, person, heavy truck, minibus, van, bus, motorcycle, taxi, light truck, and bicycle. The class distributions are as follows: Car: 185,644, Person: 37,656, Heavy truck: 12,369, Minibus: 10,777, Van: 9,251, Bus: 7,188, Motorcycle: 5,094, Taxi: 3,908, Light truck: 3,185, and Bicycle: 328.

*5.4. Cross-Junction Generalization Strategy*

Our approach is a scalable dataset expansion strategy where the training dataset constantly grows as new intersections are integrated into the system, rather than training individual models for each junction. When a new junction is added, image samples are collected under various conditions then they are annotated and included into the existing training dataset for model fine-tuning. This continuous learning framework ensures that our unified model improves generalization capabilities across diverse traffic scenarios, junction geometries, and environmental conditions while maintaining computational efficiency through a single model architecture that reduces inference time and memory requirements compared to junction-specific approaches. The methodology leverages accumulated knowledge from all previous junctions to enhance performance on new intersections, eliminates maintenance complexity of managing multiple model versions, and enables cross-junction learning where features from one intersection improve detection at geometrically similar locations, contrasting with alternative per-junction training approaches that suffer from resource intensiveness, limited generalization, exponential storage requirements, and cold start problems requiring extensive data collection for each new intersection.

*5.5. Data Splitting*

The dataset was divided into three subsets: training, validation, and testing with a split ratio of 90%, 5%, and 5%, respectively. Given the large size of the dataset, adopting a 90/5/5 split ensures that the model benefits from extensive training data while still preserving sufficient validation and testing subsets for reliable performance assessment. The training set (90%) was used to optimize the model parameters through supervised learning. The validation set (5%) was employed during training to fine-tune hyperparameters and assess the model's generalization performance, preventing overfitting. The testing set (5%) was reserved for evaluating the final model's performance on unseen data, ensuring an unbiased assessment of detection accuracy and robustness.

*5.6. Addressing Class Imbalance*

A critical challenge in our dataset was the significant class imbalance, with the majority class (car) accounting for approximately 67% of all labels, while the minority class (bicycle) represented less than 0.2%. Without intervention, such imbalances could lead to biased model predictions, favoring the majority class while underperforming on underrepresented ones. To address this, we developed a data augmentation pipeline that both increases the representation of minority classes and prevents further skewing of the dataset. The key steps include:

**Class Ratio Calculation:** For each class, we calculated a ratio based on the most frequent class (car). This ratio determined the number of additional samples needed to balance each class relative to the majority class.

**Augmentation of Minority Classes:** We generated new samples for underrepresented classes by duplicating randomly selected images and applying a variety of transformations. These transformations included:
- Adding random noise.
- Modifying brightness, contrast, or color balance.
- Introducing minor occlusions or blurring to simulate real-world conditions.

The following techniques are used in Augmentation:
- `iaa.GammaContrast((0.5,2.0))` – Adjusts contrast using a gamma function.
- `iaa.AdditiveGaussianNoise(scale=(10, 25))` – Adds random Gaussian noise.
- `iaa.AddToHueAndSaturation((-20, 20))` – Alters hue and saturation values.
- `iaa.AllChannelsCLAHE(clip_limit=(1, 4))` – Applies CLAHE (Contrast Limited Adaptive Histogram Equalization).
- `iaa.HistogramEqualization()` – Histogram equalization for better contrast.

**Exclusion of Majority Class Data:** To prevent further exacerbation of class imbalance, no additional samples were generated for the majority class (car). This deliberate exclusion ensured that the augmentation process focused solely on leveling the representation of minority classes without artificially inflating the dominant class.

Following this augmentation process, the dataset expanded to 108,000 images, with the majority class ratio reduced from 67% to 40% and minority class ratios increased from 0.2% to approximately 5%. To prevent overfitting, we deliberately limited image duplication rather than fully equalizing class distributions. Excessive replication of identical samples could compromise the model's ability to generalize to unseen data.

**Benefits of the Augmentation Strategy:** This augmentation strategy not only balanced the dataset but also enhanced the model's robustness by introducing variability within the minority classes. By diversifying the visual appearance of the augmented samples, the model learned to identify and generalize features more effectively across all classes. Additionally, the exclusion of the majority class from augmentation avoided overfitting to the most frequent class and maintained the overall balance of the dataset.

The precision–recall curves shown in Figure 10 illustrate the detection performance of the VD-OffLine-v2 system for selected vehicle categories and the aggregate model performance (ALL_CLASSES), evaluated at an IoU threshold of 0.5. The CenterNet ResNet-50 V1 FPN model, trained on 108,000 labeled samples, achieves consistently high precision values exceeding 0.9 across most recall ranges, indicating strong classification accuracy with minimal false positives. Common categories such as car, bus, minibus, van, taxi, and light truck exhibit smooth near-ideal curves approaching the upper-right corner, confirming robust and stable detection across diverse traffic and weather conditions.

Classes with fewer training examples, such as bicycle, motorcycle and person, show minor fluctuations due to class imbalance yet still maintain satisfactory precision levels. The combined ALL_CLASSES curve remains above 0.9 precision until approximately 0.85 recall at IoU = 0.5, confirming stable detection behavior under deployment-relevant overlap thresholds. Overall, these results demonstrate that the developed system achieves high detection reliability and generalization capability, supporting its applicability for real-world traffic management and analysis tasks.

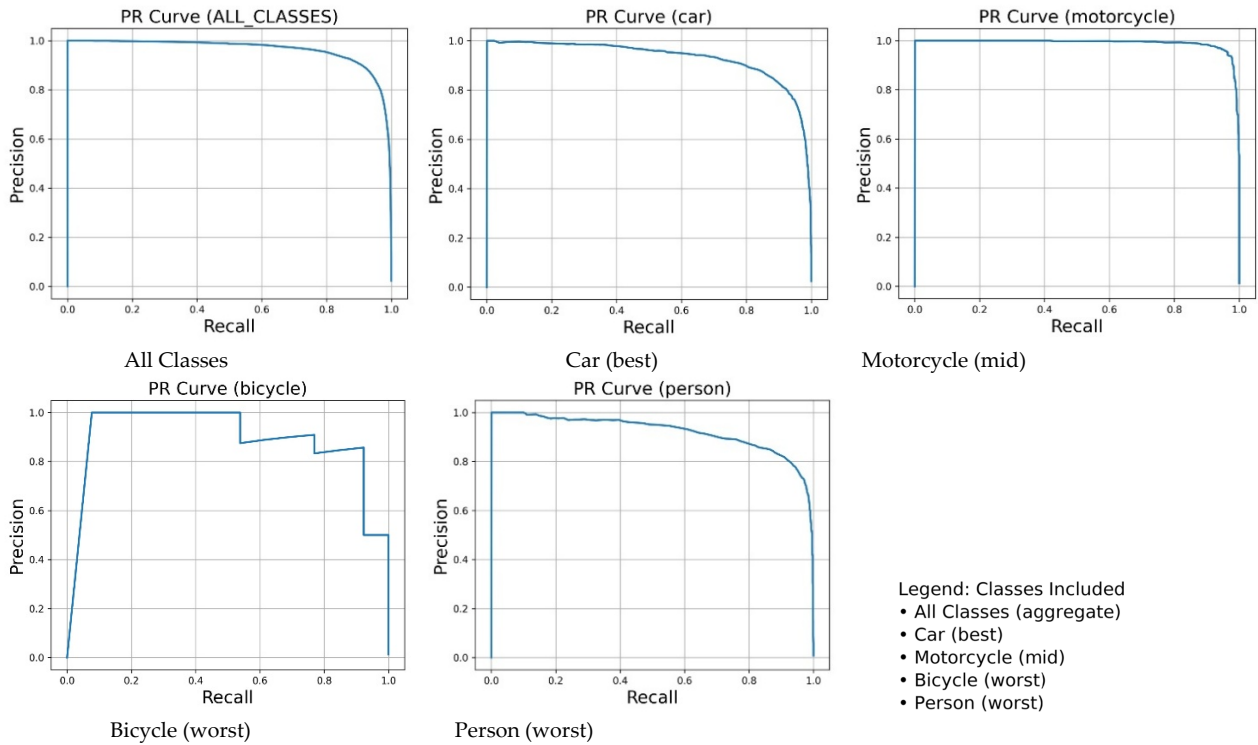

**Figure 10.** Precision–Recall (PR) curves for selected classes and the aggregate model performance.

*5.7. Model Training*

After completing data preprocessing, the model training process proceeded. For the initial and second versions of our system, we utilized the TensorFlow Object Detection API version 1.13. During these stages, we trained the SSD Inception v2 model. This model was chosen for its balance of speed and accuracy, which aligned with our initial requirement for fast detection at road junctions. As the project progressed, the need for rapid detection diminished, and greater emphasis was placed on detection accuracy. Consequently, the Faster R-CNN Inception v2 model, which provides significantly higher detection accuracy but is slower, was trained. The performance metrics for both models are summarized in Table 2.

**Transition to TensorFlow 2.14 and Advanced Models:** With the development of the third version of our system, the goal shifted toward achieving both higher detection accuracy and faster inference speeds. This led to the adoption of the TensorFlow Object Detection API version 2.14. Using this updated framework, we trained the CenterNet ResNet-50 V1 FPN (512x512) model. This model architecture offers superior accuracy compared to Faster R-CNN while maintaining faster inference times, making it an ideal choice for detecting vehicles at junctions with greater efficiency and precision.

## 6. VD Traffic Management and Analysis Functions

Increasing urbanization and rapid growth in vehicle populations require intelligent transportation infrastructure management systems to address today's mobility challenges. The effective analysis and management of traffic flow, coupled with optimal utilization of existing road infrastructure, play an essential role in sustainable urban transportation planning. Modern traffic management systems rely fundamentally on accurate, peak-time data collection to enable informed decision-making processes.

Critical traffic parameters include vehicle counting, average speed measurements, queue length estimation, and road occupancy rates. Vehicle counting, typically performed through virtual loop detection systems, quantifies the number of vehicles passing through designated detection zones within specified time intervals, providing essential data for traffic flow analysis and capacity planning. Average speed calculations serve as key inputs for dynamic speed limit optimization and congestion detection algorithms. Queue length measurements are particularly crucial for evaluating intersection performance, signal timing optimization, and overall network efficiency assessment. These integrated data collection methodologies form the foundation for intelligent traffic management systems that can adapt to real-time conditions and optimize traffic flow patterns. The traffic management and analysis functions are grouped into 12 main functions and shown in Table 3.

Table 3. The traffic management and analysis functions provided by Video Detector and similar systems.

| Functions | VD-RT | VD-Ov1 | VD-Ov2 | GoodVision [32] | MioVision [34] | DataFromSky [33] | Traffic Sensor [26] | VB-VCF [35] |
|---|---|---|---|---|---|---|---|---|
| **1. Basic Vehicle Operations** virtual loop, presence/absence, counting, average speed, tracking, queuing, average occupancy two-way traffic analysis | ● | ● | ● | ● | ● | ● | ● | ◐ |
| **2. Real-Time Processing** Live video processing, RT detection | ● | ○ | ○ | ● | ● | ● | ● | ◐ |
| **3. Vehicle Classification** Vehicle Classes: | 6 | 10 | 10 | 8 | 15+ | 20 | 7 | Basic |
| Classification Quality: *multiple vehicle types detection and classification* | ● | ● | ● | ● | ● | ● | ● | ◐ |
| **4. Event and Violation Detection** Stopped vehicles, jaywalkers, red light runners, wrong-way drivers, incidents, speed violations | ○ | ○ | ◐ | ● | ● | ● | ○ | ○ |
| **5. Advanced Traffic Analysis** Vehicle occupancy calculation, traffic density analysis, flow optimization, congestion prediction, peak hour analysis | ◐ | ◐ | ● | ● | ● | ● | ◐ | ○ |
| **6. Data Integration and Simulation** OD matrix creation, SUMO/VISSIM integration, traffic model calibration, historical data export | ○ | ● | ● | ● | ● | ● | ○ | ◐ |
| **7. Map Integration** GIS integration, GPS mapping, Google Maps, HD maps, real-time map updates | ○ | ● | ● | ● | ● | ● | ○ | ○ |
| **8. User Interface and Reporting** Customizable reports, dashboard visualization, selected direction analysis, automated reporting | ◐ | ◐ | ● | ● | ● | ● | ○ | ○ |
| **9. System Integration** Cloud integration, third-party software, IoT connectivity | ○ | ● | ● | ● | ● | ● | ○ | ○ |
| **10. Deployment and Scalability** On-premise, cloud, hybrid deployment, multisite management | ● | ● | ● | ● | ● | ● | ◐ | ○ |
| **11. Weather and Environment Adaptation** Poor weather conditions, night vision, low light, various lighting conditions, resolution adaptability | ● | ● | ● | ● | ● | ● | ● | ○ |
| **12. Cost and Customization** Cost effectiveness | High | High | High | Medium | Low | Medium | High | High |
| Customization level | High | High | V. High | Medium | Medium | Medium | High | Low |

● : provided  ○ : not provided  ◐ : partially provided
**VD-RT:** Video Detector-Real Time, **VD-Ov1:** Video Detector OffLine-v1, **VD-Ov2:** Video Detector OffLine-v2

## 7. Discussion

Unlike traditional loop detector systems that require expensive installation, road excavation, and continuous maintenance with limited traffic information capabilities, our approach eliminates physical infrastructure needs and provides comprehensive traffic analytics including virtual loop counting, vehicle classification (6 and 10 classes across versions), real-time processing, queuing analysis, and advanced event detection.

Table 3 provides a comparative view of the functional capabilities of the VD system alongside several established commercial video-based traffic analysis platforms. Overall, the VD framework offers a solid set of core functionalities—including virtual loop operations, vehicle counting, and optional real-time processing—while maintaining relatively low implementation and operational costs compared to subscription-based alternatives such as GoodVision or DataFromSky. While mature commercial systems provide broader vehicle taxonomies and advanced violation detection features, the VD framework prioritizes modularity, transparent analytics, and integration with existing infrastructure. This design choice enables municipalities to deploy targeted traffic monitoring and analysis capabilities without reliance on proprietary hardware ecosystems or long-term subscription commitments.

The VD system has operational and economic advantages compared to existing commercial and academic solutions in the traffic management domain. Our comparative evaluation given in Table 3 against leading commercial systems including GoodVision (8 classes), MioVision (15+ classes), DataFromSky (20 classes), and TrafficSensor (7 classes) reveals that our Video Detector system achieves superior cost-effectiveness and customization flexibility while maintaining comprehensive functionality.

Our benchmark tests show that VD-RT delivers processing speeds of 20.7 FPS on HD quality images with 3-4 video camera processing capability, while offering superior weather adaptability across various lighting conditions and deployment flexibility through cloud, on-premise, and hybrid architectures. Standard detection evaluation is reported using mAP@0.5 and precision–recall curves (IoU=0.5) to ensure comparability with prior object detection studies and transparent reporting of deployment-oriented performance.

Overall, the Video Detector system demonstrates that a unified architecture combining real-time intersection control with offline analytical capabilities can address core traffic monitoring requirements using vision-based sensing. By separating operational control from high-precision post-processing, the framework accommodates diverse deployment constraints while maintaining methodological transparency. Remaining gaps relative to mature commercial platforms—such as large-scale multi-camera fusion and advanced violation detection—represent clear directions for future system extensions.

To evaluate the economic feasibility of the proposed Video Detector system, a comparative cost analysis was conducted using publicly available data and recent vendor information. Commercial vision-based platforms such as GoodVision and DataFromSky operate under subscription-based pricing models, starting at approximately €299–€699 per month depending on the analysis capacity and user licenses [34], or at €2.90–€14.50 per hour for video post-processing services depending on camera type and application [33]. MioVision solutions require customized quotations that typically exceed these values due to bundled hardware and cloud-based infrastructure. In contrast, conventional inductive loop detector systems, though relatively inexpensive in unit price (€150–€500 per loop module), incur substantial installation and maintenance costs because of required pavement cutting, wiring, and frequent repairs after roadworks.

Table 4 presents an estimated 5-year total cost of ownership (TCO) comparison for different traffic detection solutions based on publicly available pricing information and typical deployment configurations. For the proposed VD system, costs are dominated by a one-time edge hardware investment, assuming reuse of existing camera infrastructure and excluding electricity and communication expenses. In contrast, inductive loop detectors incur recurring civil work and maintenance costs, while commercial vision-based platforms rely on continuous subscription or usage-based pricing, resulting in substantially higher long-term expenditures for municipalities.

Table 4. The estimated 5-year Total Cost of Ownership (TCO) comparison per intersection.

| System | Cost Components | Estimated Cost* | Notes / Assumptions |
|---|---|---|---|
| VD-RT | Edge hardware (Jetson-class), installation, minimal maintenance | €1,200 – €1,800 (one-time) | Camera cost included, no subscription fees |
| Inductive Loop Detectors | Loop hardware, road excavation, installation, recurring maintenance | €3,000 – €6,000 | Assumes 4–6 loops per junction, includes reinstallation after roadworks |
| GoodVision / Data-FromSky | Software subscription / usage-based pricing | €4,500 – €10,000 | €299–€699 per account per month or €2.9–€14.5 per account per processing hour, includes hardware |
| MioVision | Proprietary hardware + cloud services | > €8,000 | Vendor-specific quotation, bundled hardware and recurring fees |

* 5 years.

The VD system eliminates such physical infrastructure costs by leveraging existing camera networks and edge-processing hardware, resulting in a one-time setup cost and minimal ongoing expenses for calibration and data storage. This cost structure aligns with previous studies highlighting that camera-based detection systems reduce long-term operational expenses compared to embedded sensing technologies [12, 26]. Electricity and network costs were excluded consistently for all systems because these costs are largely site-dependent and comparable across solutions, and their exclusion allows clearer comparison of system-specific expenses. These findings indicate that camera-based detection combined with edge or cloud processing can substantially reduce long-term operational expenditures compared to both inductive loop detectors and subscription-based vision platforms. The presented cost structure highlights trade-offs between upfront deployment effort and recurring service fees, providing a quantitative basis for municipal decision-making.

The successful collaboration with Istanbul IT and Smart City Technologies Inc. (ISBAK) validated the transition from research concepts to deployed solutions, providing valuable technical and bureaucratic implementation insights for future intelligent transportation system deployments. Additionally, our solution demonstrates lower 5-year TCO and deployment flexibility compared to commercial competitors in the market, offering municipalities a financially better alternative to expensive proprietary systems while maintaining equivalent analytical capabilities.

Future research will focus on expanding system capabilities to meet customer demands and traffic engineering requirements. The primary development priority involves enhancing vehicle classification from 10 to 15 classes, incorporating specialized categories such as delivery trucks and emergency service vehicles.

Enhanced reporting capabilities will include specialized analytical dashboards with level-of-service (LOS), standard traffic engineering metrics that measure the quality of traffic flow conditions, calculations, intersection capacity analysis, and signal timing optimization recommendations tailored to traffic engineering workflows.

Multi-source data integration represents another key direction, expanding beyond fixed cameras to incorporate drone-based monitoring and street-level surveillance systems for corridor-wide traffic analysis. This multi-perspective approach requires developing synchronized data fusion algorithms to process multiple video streams with varying resolutions and angles while maintaining real-time performance.

Advanced analytical tools will implement traffic engineering methodologies, including multi-period origin-destination matrix generation, traffic signal warrant analysis, and automated traffic impact assessments with integration to traffic simulation software, transitioning the platform from vehicle detection to a comprehensive traffic engineering decision support system.

## 8. Conclusions

This study presents Video Detector, a vision-based vehicle detection and traffic intersection management system designed for real-time traffic monitoring and intelligent transportation systems. Experimental evaluations and field deployments demonstrate reliable vehicle detection, tracking, and traffic flow analysis under real-world conditions. By combining standardized detection metrics, transparent reporting, and practical deployment considerations, the proposed system contributes a reproducible and adaptable reference architecture for intelligent transportation applications. Future work will focus on cross-junction generalization evaluation, quantitative tracking benchmarks, and expanded multi-camera analytics.

**Author Contributions:** Writing, review and editing, M.F.Ş. and H.G.; software and models, M.F.Ş.; methodology, H.G.; project administration and funding acquisition, Ş.P.; supervision and validation, H.G. and Ş.P. All authors have read and agreed to the published version of the manuscript.

**Funding:** This research was funded by Istanbul IT and Smart City Technologies Inc. (ISBAK).

**Data Availability Statement:** The dataset developed in this study is openly available on Mendeley: Şen, Mustafa Fatih; Gümüşkaya, Haluk; Pazar, Şenol (2025), "Junction-based Vehicle Detection Dataset", doi: 10.17632/vwjg6b7kpt.1, and https://data.mendeley.com/datasets/vwjg6b7kpt/1.